\documentclass[unnumsec,webpdf,contemporary,large]{oup-authoring-template}

\graphicspath{{./}}

\usepackage{algorithm}
\usepackage{algpseudocode}

\newcommand{\btheta}{\boldsymbol{\theta}}
\newcommand{\R}{\mathbb{R}}

\makeatletter
\AtBeginDocument{%
  \def\ps@opening{%
    \def\@oddfoot{\hfil\small\sffamily\bfseries\thepage\hfil}%
    \def\@evenfoot{\hfil\small\sffamily\bfseries\thepage\hfil}%
    \let\@evenhead\relax
    \let\@oddhead\relax}%
}
\makeatother

\begin{document}

	\journaltitle{}
	\DOI{DOI HERE}
	\copyrightyear{2026}
	\pubyear{2026}
	\access{Advance Access Publication Date: Day Month Year}
	\appnotes{Original Paper}
	
	\firstpage{1}
	
	\title[StackFeat-RL]{StackFeat-RL: Reinforcement Learning over Iterative Dual-Criterion Feature Selection for Stable Biomarker Discovery}
	
	\author[1,$\ast$]{A. Yermekov}
	\author[2]{D. A. Herrera-Mart\'{i}}
	
	\authormark{Yermekov and Herrera Mart\'{i}}
	
	\address[1]{\orgname{PAfoS.AI}, \orgaddress{\state{Almaty}, \country{Kazakhstan}}}
	\address[2]{\orgdiv{CEA List}, \orgname{Universit\'{e} Grenoble Alpes}, \orgaddress{\country{France}}}
	
	\corresp[$\ast$]{Corresponding author. \href{mailto:ak.yermek@pafos.ai}{ak.yermek@pafos.ai}}
	
	\received{March}{16}{2026}
	
	\abstract{
		\textbf{Motivation:} Feature selection in high-dimensional genomic
		data ($d \gg n$) demands methods that are simultaneously accurate,
		sparse, and stable. Existing approaches either require manual
		threshold specification (mRMR, stability selection), produce
		unstable selections under data perturbation (Lasso, Boruta), or
		ignore biological structure entirely. We introduce StackFeat-RL, a
		meta-learning framework that optimises the hyperparameters of an
		iterative dual-criterion feature selection algorithm via REINFORCE
		policy gradients. The dual criterion, requiring both coefficient
		consistency and selection frequency, guards against two failure
		modes missed by single-criterion methods, while iterative
		accumulation provides convergence guarantees via the law of large
		numbers.
		\textbf{Results:} On COVID-19 miRNA data (GSE240888, 332 features)
		and three Alzheimer's disease classification tasks (GSE84422,
		13\,237 genes; Normal vs.\ Possible, Probable, and Definite AD), StackFeat-RL achieves the highest predictive
		accuracy among all evaluated methods, including ElasticNet, Boruta,
		mRMR, and stability selection, while requiring 3--4$\times$ fewer
		features. On AD Probable, StackFeat-RL significantly outperforms
		ElasticNet ($0.882$ vs.\ $0.835$; paired $t$-test, $p = 0.004$)
		with $4\times$ fewer genes. On all three AD tasks, StackFeat-RL
		significantly outperforms mRMR at matched gene counts ($p < 0.003$).
		Pathway enrichment of the Definite AD gene panel confirms
		involvement of autophagy, NRF2-mediated oxidative stress, and
		BDNF signalling. STRING network filtering reveals three
		functional modules consistent with established AD pathobiology. The method requires no manual specification of panel
		size, regularisation strength, or stopping criterion, and runs
		10--17$\times$ faster than the base algorithm.
		\textbf{Availability:} Source code at
		\url{https://github.com/pafos-ai/stackfeat-rl}.
		\textbf{Contact:} \href{mailto:ak.yermek@pafos.ai}{ak.yermek@pafos.ai}
	}
	
	
	\maketitle
	
	\section{Introduction}
	
	Gene expression profiling produces datasets where the number of
	features $d$ exceeds the number of samples $n$ by orders of
	magnitude. Clinical applications require compact, reproducible gene
	panels for downstream validation via qPCR, targeted sequencing, or
	pathway analysis. This demands methods that are accurate, sparse,
	and, crucially, stable: the selected features must be biologically interpretable and robust to
	data perturbation.
	
	Achieving all three simultaneously is the central challenge. When $d \gg n$, many feature
	subsets achieve similar predictive performance, and small data
	perturbations can produce entirely different selections. Correlated
	features exacerbate this: L1-penalised methods arbitrarily select
	one feature from a correlated group, and the choice shifts with
	minor changes in the data. A biomarker panel that changes substantially depending on which samples are included offers little confidence for clinical translation.
	
	\textbf{Filter methods} such as minimum Redundancy Maximum Relevance
	(mRMR; \cite{peng2005feature}) rank features by mutual information
	while penalising inter-feature redundancy. The ranking is fast and
	model-agnostic, but requires a manually specified cutoff $k$, an
	arbitrary design choice that shifts the burden from algorithm to user.
	Furthermore, mutual information estimates degrade in high-dimensional
	settings with limited samples, and filter methods ignore feature
	interactions that only emerge under a predictive model.
	
	\textbf{Boruta} \cite{kursa2010feature} wraps a Random Forest to
	determine feature relevance via comparison against shadow (permuted)
	features. While fully automated, it inherits the base learner's
	limitations: Random Forests handle correlated features poorly, split
	importance arbitrarily between redundant predictors, and lack any
	mechanism to incorporate external biological knowledge. Boruta
	typically produces large panels (100--150 genes in our experiments),
	limiting clinical applicability.
	
	\textbf{Stability selection} \cite{meinshausen2010stability}
	subsamples data repeatedly and runs Lasso at multiple regularisation
	strengths, selecting features that are consistently chosen across
	subsamples. It thresholds on selection frequency alone, which means
	a feature selected in every subsample but with alternating coefficient
	sign (evidence of instability, not reliability) would pass the
	threshold. Furthermore, while individual LASSO fits are predictive, the aggregation step only counts how often each feature appears, without evaluating the final panel's predictive performance or controlling its size.
	
	\textbf{ElasticNet} \cite{zou2005regularization} with
	cross-validated regularisation (ElasticNetCV) provides a strong
	embedded baseline. However, a single fit produces a one-shot
	selection that depends on the particular data partition. The L2
	component retains groups of correlated variables rather than
	picking a representative, yielding panels of 170--200 genes on
	high-dimensional data, well beyond what is practical for clinical validation. Running
	ElasticNet on a different train/test split can produce a
	substantially different gene list, with no principled way to
	identify which selections are reliable.
	
	We previously introduced StackFeat \cite{yermekov2026stackfeat}, an
	iterative algorithm that accumulates two statistics across multiple reshuffled data partitions: signed coefficients (measuring effect strength and
	direction) and selection frequencies (estimating selection
	probability). Only features ranking highly under \emph{both} criteria
	are retained. This dual requirement guards against two failure modes
	that single-criterion methods miss: (1)~sign-inconsistent features,
	selected frequently but with cancelling coefficients; and
	(2)~infrequent-but-consistent features, overshadowed by correlated
	alternatives. Iterative accumulation provides convergence guarantees
	via the law of large numbers: normalised statistics converge to
	population-level importance measures as iterations increase.
	
	StackFeat requires nested cross-validation at every iteration
	(ElasticNetCV per fold per iteration), making it computationally
	prohibitive on high-dimensional data ($d > 10\,000$). Furthermore,
	its key hyperparameter, the feature retention fraction
	m\_frac, is fixed at $1/4$, a heuristic that may be suboptimal
	for a given dataset.
	
	In this work, we present \textbf{StackFeat-RL}, which recasts the
	hyperparameter optimisation of StackFeat as a reinforcement learning
	problem. A REINFORCE policy \cite{williams1992simple} learns the
	feature retention fraction and per-gene penalty modulation from data,
	while a single ElasticNetCV call per evaluation fold replaces the
	expensive nested regularisation search, reducing computation by
	10--17$\times$. The framework accepts biological priors (e.g.,
	STRING protein interaction networks) through the policy's state
	representation, providing a principled interface for incorporating
	domain knowledge. StackFeat-RL requires no manual specification of
	panel size, regularisation strength, or stopping criterion: all are
	learned or derived from the data.
	
	\section{Methods}
	
	\subsection{Notation and state representation}
	
	Let $\mathbf{X} \in \R^{n \times p}$ be the expression matrix,
	$\mathbf{y} \in \R^n$ the binary response,
	$\hat{\boldsymbol{\beta}}^{(t,f)} \in \R^p$ the ElasticNet
	coefficient vector (signed regression weights) at iteration $t$,
	fold $f$, and
	$\mathbf{M} \in \R^{p \times p}$ the STRING protein interaction
	matrix \cite{szklarczyk2023string} (v12, combined score $> 700$,
	corresponding to high-confidence interactions on STRING's
	0--1000 scale).
	
	At each iteration $t$, per-gene accumulation statistics from
	iterations $1, \ldots, t{-}1$ with $k$ folds define the state.
	For each gene $i$:
	\begin{align}
		\hat{p}_{i,t} &= \frac{1}{(t{-}1) k}
		\textstyle\sum_{\tau=1}^{t-1} \sum_{f=1}^{k}
		\mathbf{1}[\hat{\beta}_i^{(\tau,f)} \neq 0]
		\label{eq:phat} \\
		|\hat{\mu}_{i,t}| &= \left|\frac{1}{(t{-}1) k}
		\textstyle\sum_{\tau=1}^{t-1} \sum_{f=1}^{k}
		\hat{\beta}_i^{(\tau,f)}\right|
		\label{eq:muhat} \\
		n_{i,t} &= \frac{1}{|\mathcal{S}_{t-1}|}
		\textstyle\sum_{j \in \mathcal{S}_{t-1}} M_{ij}
		\label{eq:nit} \\
		d_{i,t} &= \frac{1}{|\mathcal{S}_{t-1}|}
		\textstyle\sum_{j \in \mathcal{S}_{t-1}} \psi_{ij}
		\label{eq:dit}
	\end{align}
	where $\hat{p}_{i,t}$ is the normalised selection frequency
	(fraction of fold-iterations in which gene $i$ received a nonzero
	coefficient), $|\hat{\mu}_{i,t}|$ the normalised absolute mean
	of the signed ElasticNet coefficients (a positive coefficient
	indicates the gene is upregulated in the disease group relative
	to controls, negative indicates downregulation; $|\hat{\mu}|$ is
	high when the direction is consistent across folds and low when
	positive and negative coefficients cancel), $n_{i,t}$ the mean STRING interaction with the
	previously selected gene set $\mathcal{S}_{t-1}$, and $d_{i,t}$
	the mean co-selection frequency with $\mathcal{S}_{t-1}$. The co-selection frequency
	$\psi_{ij} = (Tk)^{-1} \sum_{t,f}
	\mathbf{1}[\hat{\beta}_i^{(t,f)} \neq 0 \wedge
	\hat{\beta}_j^{(t,f)} \neq 0]$
	measures how often genes $i$ and $j$ were simultaneously selected.These
	normalised estimators keep all state features bounded regardless
	of iteration count, preventing sigmoid saturation.
	
	\subsection{Dual-criterion selection and convergence}
	
	The dual criterion requires both coefficient consistency \emph{and}
	selection frequency. At iteration $t$:
	\begin{equation}
		S^{(t)} = \underbrace{\{j : |w_j^{(t)}| \in \text{top } m\}}_{S_w}
		\cap
		\underbrace{\{j : c_j^{(t)} \in \text{top } m\}}_{S_c}
	\end{equation}
	where $w_j^{(t)} = \sum_{\tau=1}^{t}\sum_{f=1}^{k}\hat{\beta}_j^{(\tau,f)}$
	and $c_j^{(t)} = \sum_{\tau=1}^{t}\sum_{f=1}^{k}\mathbf{1}[\hat{\beta}_j^{(\tau,f)} \neq 0]$.
	
	This guards against two failure modes:
	
	\begin{center}
		\begin{tabular}{lccc}
			\toprule
			Feature type & $|w_j|$ & $c_j$ & Outcome \\
			\midrule
			True signal & high & high & Selected \\
			Noise & low & low & Rejected \\
			Sign-inconsistent & low & high & Rejected by $S_w$ \\
			Infrequent but consistent & high & low & Rejected by $S_c$ \\
			\botrule
		\end{tabular}
	\end{center}
	
	\textbf{Convergence guarantee.} Since iterations use independent
	random seeds for fold assignment, by the law of large numbers (LLN):
	\begin{equation}
		\frac{w_j^{(T)}}{T \cdot k} \to \bar{\mu}_j, \qquad
		\frac{c_j^{(T)}}{T \cdot k} \to p_j
	\end{equation}
	where $\bar{\mu}_j = \mathbb{E}[\hat{\beta}_j]$ is the expected
	ElasticNet coefficient and
	$p_j = P(\hat{\beta}_j \neq 0)$ is the selection probability,
	both defined over the joint distribution of data partitions
	and fold assignments. As iterations accumulate,
	the selected set converges to the population-level optimal panel.
	This guarantee carries over to StackFeat-RL: the RL policy modulates
	penalties but the dual-criterion accumulation mechanism is unchanged,
	so the LLN guarantee applies independently within each episode's
	$T \cdot k$ fold-iterations. Multi-episode training improves the
	policy parameters (and thus the quality of the resulting panel),
	while the convergence of the accumulation statistics is governed by
	the same LLN as in StackFeat.
	
	\subsection{Policy}
	
	The policy $\pi_{\btheta}$ is parameterised by
	$\btheta = (\theta_1, \ldots, \theta_5) \in \R^5$. The first four
	components modulate per-gene penalties; $\theta_5$ controls the
	feature retention fraction.
	
	For gene $i$ at iteration $t$, the policy score is:
	\begin{equation}
		z_{i,t} = \theta_1 \hat{p}_{i,t}
		+ \theta_2 |\hat{\mu}_{i,t}|
		+ \theta_3 n_{i,t}
		+ \theta_4 d_{i,t}
		\label{eq:score}
	\end{equation}
	The gene-specific penalty is
	$\lambda_{i,t} = \alpha \cdot \sigma(z_{i,t})$,
	where $\sigma(\cdot)$ is the logistic sigmoid and $\alpha$ the
	regularisation strength determined by ElasticNetCV (see below).
	
	The feature retention fraction replaces a hand-tuned value:
	\begin{align}
		\text{m\_frac} &= 0.25 + 0.65 \cdot \sigma(\theta_5)
		& &\in [0.25, 0.90] \label{eq:mfrac}
	\end{align}
	
	\textbf{Regularisation.} The global regularisation strength
	$\alpha$ is determined by a single ElasticNetCV call (5-fold, 100
	alphas) on the outer training data before the episode loop, and
	held fixed for all episodes. This reduces computation by
	10--17$\times$ compared to running ElasticNetCV at every iteration.
	
	The feature retention fraction m\_frac replaces the fixed $1/4$
	heuristic of StackFeat: $m = \max(3, \lfloor
	\text{mean(genes\_per\_fold)} \cdot \text{m\_frac} \rfloor)$.
	Intuitively, m\_frac controls how aggressively the dual criterion
	filters: a small value (e.g., 0.25) considers only the top quartile
	under each criterion, risking premature exclusion of informative
	genes that rank slightly below the cutoff on one criterion.
	A larger value widens the pool, letting the intersection do
	more of the filtering. The policy learns the appropriate trade-off
	from data.
	
	$\btheta$ is initialised to $\mathbf{0}$, yielding uniform
	penalties and $\text{m\_frac} = 0.25 + 0.65 \cdot \sigma(0) = 0.575$.
	
	\subsection{Algorithm}
	
	\begin{algorithm}[!t]
		\caption{StackFeat-RL (one outer fold)}
		\label{alg:stackfeat-rl}
		\footnotesize
		\begin{algorithmic}[1]
			\Require $(\mathbf{X}_{\text{tr}}, \mathbf{y}_{\text{tr}})$,
			$\mathbf{M}$, folds $k$, episodes $K$,
			learning rate $\alpha_\pi$, baseline decay $\gamma$,
			tolerance $\varepsilon$, sparsity weight $\lambda_s$
			\State $\alpha \gets \text{ElasticNetCV}(\mathbf{X}_{\text{tr}},
			\mathbf{y}_{\text{tr}}).\alpha\_$
			\Comment{ElasticNetCV}
			\State $\btheta \gets \mathbf{0}$, \; $b \gets 0$
			\For{episode $e = 1, \ldots, K$}
			\State Reset $\mathbf{w}, \mathbf{c},
			\mathcal{G}_e, \boldsymbol{\psi} \gets \mathbf{0}$
			\State Compute m\_frac from $\btheta$
			via Eq.~\ref{eq:mfrac}
			\Statex \hspace{\algorithmicindent}\textit{--- Warmup ($t{=}1$):
				uniform penalties, no gradient ---}
			\State Run $k$-fold ElasticNet($\alpha$);
			update $\mathbf{w}, \mathbf{c}$
			\Statex \hspace{\algorithmicindent}\textit{--- Policy loop
				($t = 2, \ldots, T$) ---}
			\For{$t = 2, \ldots, T_{\max}$}
			\State Compute $\hat{p}_{i,t}, |\hat{\mu}_{i,t}|, n_{i,t},
			d_{i,t}$ $\forall\, i$
			\State $z_{i,t} \gets \theta_1 \hat{p}_{i,t}
			+ \theta_2 |\hat{\mu}_{i,t}|
			+ \theta_3 n_{i,t}
			+ \theta_4 d_{i,t}$
			\State $\lambda_{i,t} \gets \alpha \cdot \sigma(z_{i,t})$
			$\forall\, i$
			\State Run $k$-fold ElasticNet with
			$\boldsymbol{\lambda}_t$;
			update $\mathbf{w}, \mathbf{c},
			\boldsymbol{\psi}$
			\State $\mathcal{G}_e \mathrel{+}=
			\sum_i (\mathbf{1}[i \notin S_t]
			- \sigma(z_{i,t})) \mathbf{s}_{t,i}$
			\State \textbf{if} 2 consecutive AUC diffs
			$< \varepsilon$ \textbf{then break}
			\EndFor
			\State $\mathcal{G}_e \gets \mathcal{G}_e \,/\,
			(p \cdot (T{-}1))$
			\Comment{Normalise gradient}
			\State $m \gets \max(3, \lfloor \text{mean\_genes}
			\cdot \text{m\_frac} \rfloor)$
			\State $S^* \gets \{j : |w_j| \in \text{top } m\}
			\cap \{j : c_j \in \text{top } m\}$
			\State $R_e \gets \text{AUC}(S^*)
			- \lambda_s |S^*|$
			\State $b \gets \gamma b + (1{-}\gamma) R_e$;
			\; $\btheta \gets \btheta
			+ \alpha_\pi (R_e - b) \mathcal{G}_e$
			\EndFor
			\State \Return $S^*$, $\btheta$, $\boldsymbol{\psi}$
		\end{algorithmic}
	\end{algorithm}
	
	\subsection{Gradient derivation}
	
	The REINFORCE objective is
	$J(\btheta) = \mathbb{E}_{\tau \sim \pi_{\btheta}}[R(\tau)]$.
	By the log-derivative trick:
	\begin{equation}
		\nabla_{\btheta} J = \mathbb{E}\!\left[
		(R - b) \textstyle\sum_{t=2}^{T}
		\nabla_{\btheta} \log P(\mathcal{S}_t \mid
		\mathbf{s}_t, \btheta) \right]
		\label{eq:reinforce}
	\end{equation}
	Modelling gene selections as independent Bernoulli outcomes with
	exclusion probability $\pi_i = \sigma(z_i)$:
	\begin{equation}
		\nabla_{\btheta} \log P(\mathcal{S}_t \mid
		\mathbf{s}_t, \btheta)
		= \textstyle\sum_{i=1}^{p}
		(\mathbf{1}[i \notin \mathcal{S}_t] - \sigma(z_{i,t}))
		\mathbf{s}_{t,i}
		\label{eq:gradient}
	\end{equation}
	The accumulated gradient is normalised by $p \cdot (T{-}1)$ for
	scale invariance across datasets of different dimensionality.
	
	\subsection{Posterior network}
	\label{sec:posterior}
	
	The posterior network $M^*_{ij} = M_{ij} \cdot \psi_{ij}$ retains a
	STRING edge only if both endpoints were repeatedly co-selected,
	filtering the general interactome to disease-relevant interactions.
	
	\subsection{Computational complexity}
	
	\begin{table}[!t]
		\begin{center}
			\begin{minipage}{\columnwidth}
				\caption{Total elastic net fits per outer fold.\label{tab:complexity}}%
				\begin{tabular*}{\columnwidth}{@{\extracolsep\fill}lrl@{\extracolsep\fill}}
					\toprule
					Method & Formula & Fits \\
					\midrule
					ElasticNet (nested) & $A \cdot k_{\text{cv}}$ & $\sim$500 \\
					SF-RL (nested) & $A k_{\text{cv}} + K T k$ & $\sim$725 \\
					Stab.\ Sel. & $|\Lambda| \cdot B$ & $\sim$2\,500 \\
					StackFeat (nested) & $T k A k_{\text{cv}}$ & $\sim$12\,500 \\
					\botrule
				\end{tabular*}
				\begin{tablenotes}
					\item $A = 100$ alphas, $k_{\text{cv}} = 5$ CV folds,
					$K = 15$ episodes, $T \approx 3$ iterations, $k = 5$ inner folds,
					$|\Lambda| = 25$ regularisation strengths, $B = 100$ subsamples.
					SF-RL: 500 (ElasticNetCV) $+$ $15 \times 3 \times 5 = 225$
					single-fit calls $= 725$ per outer fold. On 13\,237 features,
					this is the difference between computationally feasible (SF-RL)
					and prohibitive (StackFeat). Furthermore, REINFORCE episodes
					are independent and can be run in parallel, reducing wall-clock
					time by up to $K\times$ on multi-core hardware.
				\end{tablenotes}
			\end{minipage}
		\end{center}
	\end{table}
	
	\section{Experimental Setup}
	
	\subsection{Datasets}
	
	\textbf{COVID-19 miRNA} (GSE240888): 122 serum samples (60 COVID-19,
	62 control), 332 miRNA features. This low-dimensional dataset
	demonstrates graceful degradation when adaptive penalties are
	unnecessary, and enables comparison with the 5-miRNA signature from
	\cite{yermekov2026stackfeat}.
	
	\textbf{Alzheimer's disease} (GSE84422 \cite{liang2022gse84422}, GPL96): 951 samples from 19
	brain regions, 13\,237 genes (Affymetrix HG-U133A, RMA normalised).
	Four diagnostic categories: Normal (214), Possible AD (229),
	Probable AD (180), Definite AD (328). Three binary tasks: Normal vs.\
	Possible (443 samples), Normal vs.\ Probable (394 samples), Normal
	vs.\ Definite (542 samples). STRING v12 protein--protein interaction network (combined score
	$> 700$; 917\,347 interactions mapped to the 13\,237 measured genes).
	
	\subsection{Baselines}
	
	All methods evaluated under identical 10-fold nested CV with the
	same outer fold splits and a logistic regression classifier trained
	on selected features.
	
	(1)~\textbf{ElasticNet}: ElasticNetCV-selected $\alpha$ (5-fold, 100
	alphas). A strong, properly tuned embedded baseline.
	(2)~\textbf{Boruta}: Random Forest wrapper with
	\texttt{class\_weight=balanced}.
	(3)~\textbf{mRMR}: Maximum relevance, minimum redundancy filter with
	$k$ matched to StackFeat-RL's average gene count per task (COVID:
	$k = 9$; AD Possible: $k = 48$; AD Probable: $k = 42$;
	AD Definite: $k = 56$) for fair comparison at equal gene budgets.
	(4)~\textbf{Stability selection} (COVID-19 only): Standard
	Meinshausen \& B\"{u}hlmann algorithm with 25 regularisation
	strengths ($\Lambda = 10^{-3}$--$10^{-1}$), 100 subsamples without
	replacement, selection threshold $\pi_{\text{thr}} = 0.9$.
	Computationally prohibitive on the AD dataset (2\,500 Lasso fits
	per fold $\times$ 13\,237 features).
	(5)~\textbf{StackFeat} (COVID-19 only): Base dual-criterion
	algorithm ($\btheta = \mathbf{0}$, m\_frac $= 0.25$, ElasticNetCV
	per fold). Omitted on AD due to computational cost
	($\sim$12\,500 fits per outer fold vs.\ $\sim$725 for SF-RL).
	
	\subsection{Hyperparameters}
	
	StackFeat-RL: learning rate $\alpha_\pi = 0.5$, baseline decay
	$\gamma = 0.9$, convergence tolerance $\varepsilon = 0.02$, sparsity
	weight $\lambda_s = 0.001$, $K = 15$ episodes, $k = 5$ inner folds,
	$F = 10$ outer folds, minimum 3 genes. Gradient normalised by
	$p \cdot (T{-}1)$. Clipping: $\theta_5 \in [-4, 4]$.
	
	The reward $R = \text{AUC}(S^*) - \lambda_s |S^*|$ balances
	predictive accuracy with panel compactness; at 50 genes the
	sparsity penalty is 0.05, well below typical AUC differences
	between methods.
	
	All models implemented using scikit-learn \cite{pedregosa2011sklearn}.
	
	\section{Results}
	
	\subsection{COVID-19 miRNA}
	
	\begin{table}[!t]
		\begin{center}
			\begin{minipage}{\columnwidth}
				\caption{Nested CV results: COVID-19 miRNA (GSE240888, 10 outer folds).
					Avg genes = mean genes selected per fold.
					Cons.\ = consensus genes ($\geq 6/10$ folds).\label{tab:covid}}%
				\begin{tabular*}{\columnwidth}{@{\extracolsep\fill}lccccc@{\extracolsep\fill}}
					\toprule
					Method & Mean AUC & Std & Median & Avg genes & Cons. \\
					\midrule
					mRMR ($k{=}9$) & 0.865 & 0.125 & 0.825 & 9.0 & 3 \\
					Stab.\ Sel. & 0.875 & 0.132 & 0.875 & 46.5 & 46 \\
					StackFeat & 0.880 & 0.151 & 0.975 & 4.6 & 4 \\
					Boruta & 0.890 & 0.133 & 0.975 & 25.1 & 20 \\
					SF-RL & 0.895 & 0.126 & 0.975 & \textbf{8.8} & \textbf{8} \\
					ElasticNet & \textbf{0.906} & 0.123 & 0.975 & 20.1 & 12 \\
					\botrule
				\end{tabular*}
				\begin{tablenotes}
					\item No pairwise differences reach significance (paired $t$-test,
					$p > 0.19$), consistent with all methods achieving near-identical
					median AUC on this low-dimensional dataset. Stability selection
					selects 46.5 genes per fold despite $\pi_{\text{thr}} = 0.9$,
					illustrating threshold sensitivity on small feature spaces.
				\end{tablenotes}
			\end{minipage}
		\end{center}
	\end{table}
	
	All methods achieve similar performance on this low-dimensional
	dataset. StackFeat-RL produces an 8-gene consensus panel
	(hsa-miR-107, hsa-miR-181a-5p, hsa-miR-181b-5p, hsa-miR-150-5p,
	hsa-miR-484, hsa-miR-485-3p, hsa-miR-625-3p, hsa-miR-1185-1-3p),
	a superset of the 5-gene signature in \cite{yermekov2026stackfeat},
	with median AUC 0.975 matching ElasticNet and Boruta. ElasticNet
	achieves the highest mean (0.906) but uses $2.3\times$ more features.
	
	\begin{figure*}[!t]
		\centering
		\includegraphics[width=\textwidth]{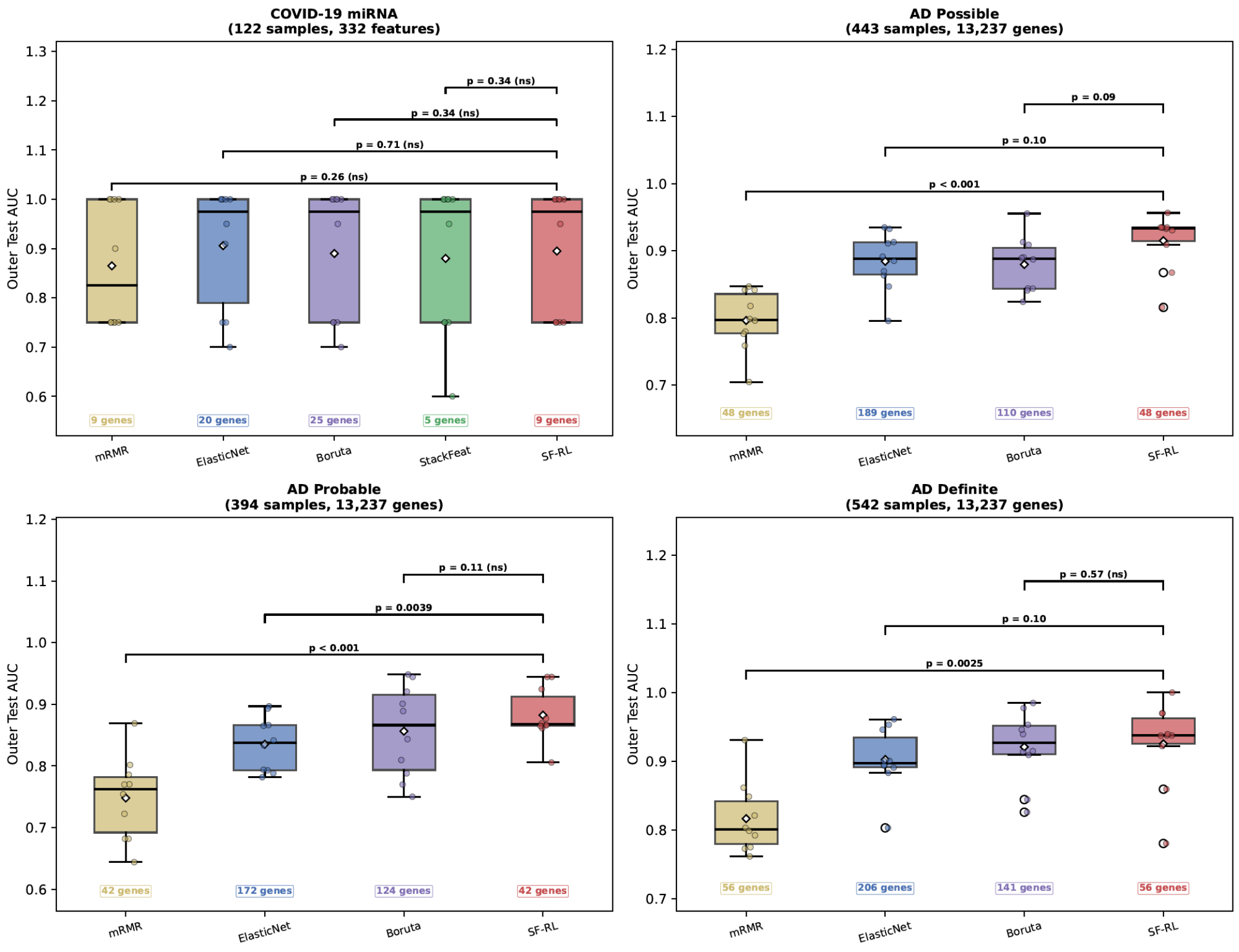}
		\caption{Nested CV performance (AUC) across all methods and tasks.
			Average gene counts shown below each box.
			Diamond: mean; horizontal line: median; dots: individual outer folds.
			Brackets: paired $t$-test $p$-values vs.\ StackFeat-RL.
			StackFeat-RL achieves the highest mean AUC on all three AD tasks
			while using 3--4$\times$ fewer genes than ElasticNet and Boruta.}
		\label{fig:boxplots}
	\end{figure*}
	
	\subsection{Alzheimer's disease}
	
	\begin{table}[!t]
		\begin{center}
			\begin{minipage}{\columnwidth}
				\caption{Nested CV results: Alzheimer's disease (GSE84422, 10 outer
					folds, 3 binary tasks). mRMR $k$ matched to SF-RL's average gene
					count per task. $p$-values: paired $t$-test vs.\ SF-RL.\label{tab:ad}}%
				\begin{tabular*}{\columnwidth}{@{\extracolsep\fill}llccccc@{\extracolsep\fill}}
					\toprule
					Task & Method & Mean & Std & Med. & Genes & $p$ \\
					\midrule
					\multicolumn{7}{@{}l}{\textit{Normal vs.\ Possible AD (443 samples)}} \\
					& mRMR ($k{=}48$) & 0.796 & 0.044 & 0.797 & 48.0 & $<$0.001 \\
					& ElasticNet & 0.884 & 0.043 & 0.888 & 188.9 & 0.096 \\
					& Boruta & 0.880 & 0.041 & 0.888 & 109.9 & 0.090 \\
					& \textbf{SF-RL} & \textbf{0.915} & 0.042 & \textbf{0.934} & \textbf{47.8} & --- \\
					\midrule
					\multicolumn{7}{@{}l}{\textit{Normal vs.\ Probable AD (394 samples)}} \\
					& mRMR ($k{=}42$) & 0.748 & 0.067 & 0.762 & 42.0 & $<$0.001 \\
					& ElasticNet & 0.835 & 0.044 & 0.837 & 172.3 & \textbf{0.004} \\
					& Boruta & 0.856 & 0.074 & 0.866 & 124.2 & 0.109 \\
					& \textbf{SF-RL} & \textbf{0.882} & 0.043 & \textbf{0.868} & \textbf{41.8} & --- \\
					\midrule
					\multicolumn{7}{@{}l}{\textit{Normal vs.\ Definite AD (542 samples)}} \\
					& mRMR ($k{=}56$) & 0.816 & 0.052 & 0.801 & 56.0 & \textbf{0.003} \\
					& ElasticNet & 0.903 & 0.045 & 0.897 & 205.8 & 0.096 \\
					& Boruta & 0.921 & 0.052 & 0.927 & 140.9 & 0.567 \\
					& \textbf{SF-RL} & \textbf{0.925} & 0.063 & \textbf{0.938} & \textbf{55.6} & --- \\
					\botrule
				\end{tabular*}
				\begin{tablenotes}
					\item Bold $p$-values: significant at $\alpha = 0.05$ (Bonferroni
					correction not applied; individual comparisons reported).
					SF-RL achieves the highest mean AUC on all three tasks with
					3--4$\times$ fewer genes than ElasticNet and Boruta.
				\end{tablenotes}
			\end{minipage}
		\end{center}
	\end{table}
	
	StackFeat-RL achieves the highest mean AUC on all three AD tasks.
	On Probable AD, SF-RL significantly outperforms ElasticNet (paired
	$t$-test, $p = 0.004$; $+4.7\%$ AUC with $4.1\times$ fewer genes:
	41.8 vs.\ 172.3). On Possible and Definite AD, the improvement
	approaches significance ($p = 0.096$) with $3.0\times$ and
	$3.7\times$ gene reduction respectively. StackFeat-RL significantly
	outperforms mRMR on all three tasks ($p < 0.003$) at matched gene
	counts, demonstrating that the advantage comes from \emph{which}
	genes are selected, not how many.
	
	Boruta achieves competitive accuracy (0.921 on Definite AD) but
	requires 140.9 genes per fold, $2.5\times$ more than SF-RL's 55.6.
	ElasticNet uses 173--206 genes per fold across tasks, producing
	verbose panels with limited clinical utility.
	
	\begin{figure*}[!t]
		\centering
		\includegraphics[width=\textwidth]{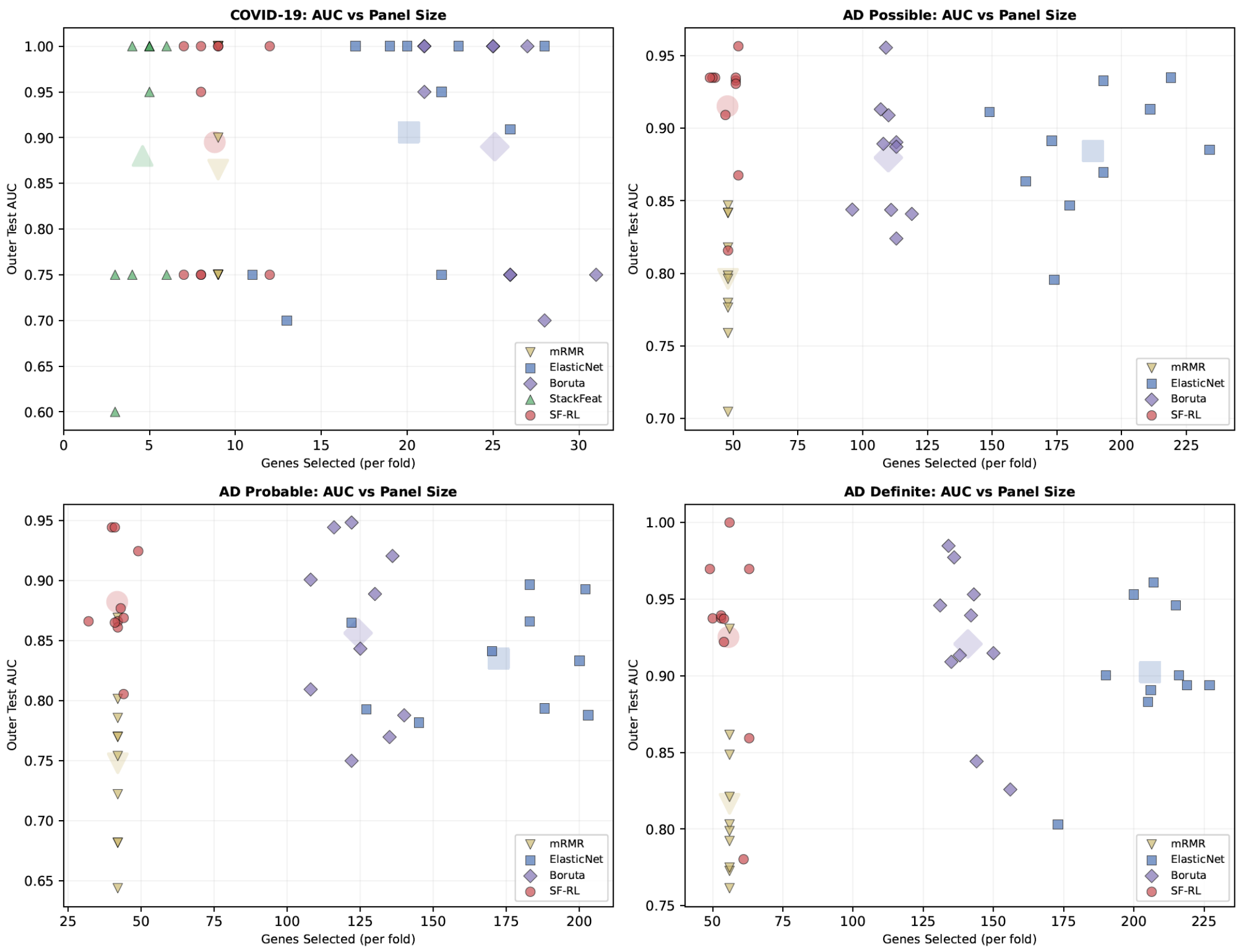}
		\caption{AUC vs.\ panel size (genes per fold). Each point represents
			one outer fold. Large translucent markers indicate method means.
			On all AD tasks, StackFeat-RL occupies the upper-left quadrant
			(high accuracy, few genes), demonstrating Pareto dominance over
			all baselines. mRMR (matched gene count) and SF-RL overlap on the
			$x$-axis but are clearly separated on AUC.}
		\label{fig:pareto}
	\end{figure*}
	
	\subsection{Learned policy parameters}
	
	\begin{table}[!t]
		\begin{center}
			\begin{minipage}{\columnwidth}
				\caption{Average learned $\btheta$ across 10 outer folds.
					m\_fr = feature retention fraction (StackFeat default:
					0.25).\label{tab:theta}}%
				\begin{tabular*}{\columnwidth}{@{\extracolsep\fill}lcccccc@{\extracolsep\fill}}
					\toprule
					Dataset & $\theta_1$ & $\theta_2$ & $\theta_3$ & $\theta_4$ & $\theta_5$ & m\_fr \\
					\midrule
					COVID-19 & $+$0.0041 & $+$0.0004 & 0.0 & 0.0 & $-$0.066 & 0.564 \\
					AD Poss. & $+$0.0019 & $+$0.0002 & $-$0.0008 & $+$0.0007 & $-$0.485 & 0.498 \\
					AD Prob. & $+$0.0017 & $+$0.0001 & $-$0.0008 & $+$0.0006 & $-$0.500 & 0.495 \\
					AD Def. & $+$0.0027 & $+$0.0002 & $-$0.0009 & $+$0.0011 & $-$0.563 & 0.486 \\
					\botrule
				\end{tabular*}
				\begin{tablenotes}
					\item $\theta_1$--$\theta_4 \approx 0$: the policy learns that
					uniform penalties are sufficient on all tested datasets. $\theta_5$
					controls m\_frac via $0.25 + 0.65 \cdot \sigma(\theta_5)$;
					learned values ($0.49$--$0.56$) roughly double the StackFeat
					default of $0.25$. $\alpha$ determined by ElasticNetCV per outer
					fold (typical values $0.02$--$0.06$).
				\end{tablenotes}
			\end{minipage}
		\end{center}
	\end{table}
	
	Per-gene parameters ($\theta_1$--$\theta_4$) remain near zero on
	all datasets. This is itself a finding: the policy learns that
	uniform penalties are already sufficient for these phenotypes, and
	no per-gene modulation is needed. The only parameter that moves meaningfully is $\theta_5$ (m\_frac),
	which the policy adjusts from the initial 0.575 to 0.486--0.564
	across datasets. Even after adjustment, m\_frac remains roughly
	double the StackFeat default of 0.25, retaining a wider candidate
	pool for dual-criterion intersection. With
	$\btheta = \mathbf{0}$, StackFeat-RL reduces
	exactly to StackFeat with optimal regularisation, so the framework
	never degrades below the base algorithm's performance.

	\subsection{Gene panel stability and overlap}
	
	Consensus gene panels are largely stage-specific: only 4 genes
	(AMFR, ENAH, HBA1, TNNT1) overlap between the 33-gene Possible AD
	panel and the 35-gene Probable AD panel (12\% Jaccard similarity).
	This suggests distinct molecular mechanisms at different disease
	stages, consistent with AD's heterogeneous progression.
	
	\subsection{Biological validation}
	
	Pathway enrichment analysis of the 80-gene Definite AD panel
	(the union of genes selected across all 10 outer folds;
	Enrichr; \cite{kuleshov2016enrichr}) revealed significant
	enrichment across multiple databases after false discovery rate
	correction:
	
	\begin{itemize}
		\item \textbf{Reactome}: Chaperone-mediated autophagy
		(adj.\ $p = 0.015$; OR $= 40.8$), autophagy
		(adj.\ $p = 0.033$; OR $= 8.6$)
		\item \textbf{WikiPathways}: NRF2 pathway
		(adj.\ $p = 0.021$; OR $= 9.7$), neuroinflammation and glutamatergic
		signalling ($p = 0.002$)
		\item \textbf{BioPlanet}: Alpha-haemoglobin stabilising enzyme
		pathway (adj.\ $p = 0.005$; OR $= 77.6$), BDNF signalling
		(adj.\ $p = 0.033$; OR $= 6.3$)
		\item \textbf{KEGG}: Hormone signalling (adj.\ $p = 0.030$),
		glycine/serine/threonine metabolism (adj.\ $p = 0.030$; OR $= 22.1$)
		\item \textbf{MSigDB Hallmark}: Hypoxia (adj.\ $p = 0.037$;
		OR $= 6.7$)
	\end{itemize}
	
	NRF2-mediated oxidative stress response, neuroinflammation,
	impaired autophagy, and cerebrovascular hypoxia are established
	mechanisms in AD pathogenesis, confirming the biological relevance
	of the selected gene panel.
	
	\textbf{STRING network analysis.} Filtering the STRING protein
	interaction network by co-selection frequency ($\psi$) reveals 9
	interactions among 9 genes, forming three biologically coherent
	modules:
	
	\begin{enumerate}
		\item \textbf{Haemoglobin/haeme biosynthesis}: HBB--HBA1
		($\psi = 1.00$), HBB--HBG1 ($\psi = 0.80$), HBB--ALAS2
		($\psi = 0.73$), HBA1--HBG1 ($\psi = 0.80$), DAO--ALAS2
		($\psi = 0.60$). Consistent with cerebrovascular oxygen transport
		dysfunction in AD.
		\item \textbf{Oxidative stress}: GSTM3--PRDX6 ($\psi = 0.67$).
		Glutathione S-transferase and peroxiredoxin, key enzymes in the
		oxidative damage response implicated in AD.
		\item \textbf{Autophagy}: ATG101--ATG4B ($\psi = 0.73$). Core
		autophagy machinery, directly supporting the Reactome enrichment
		result (chaperone-mediated autophagy, adj.\ $p = 0.015$).
	\end{enumerate}
	
	These modules provide converging evidence from two independent
	analyses (pathway enrichment and network topology) for the same
	biological mechanisms, strengthening the case that StackFeat-RL
	selects functionally coherent gene panels.
	
	\begin{figure}[!t]
		\centering
		\includegraphics[width=\columnwidth]{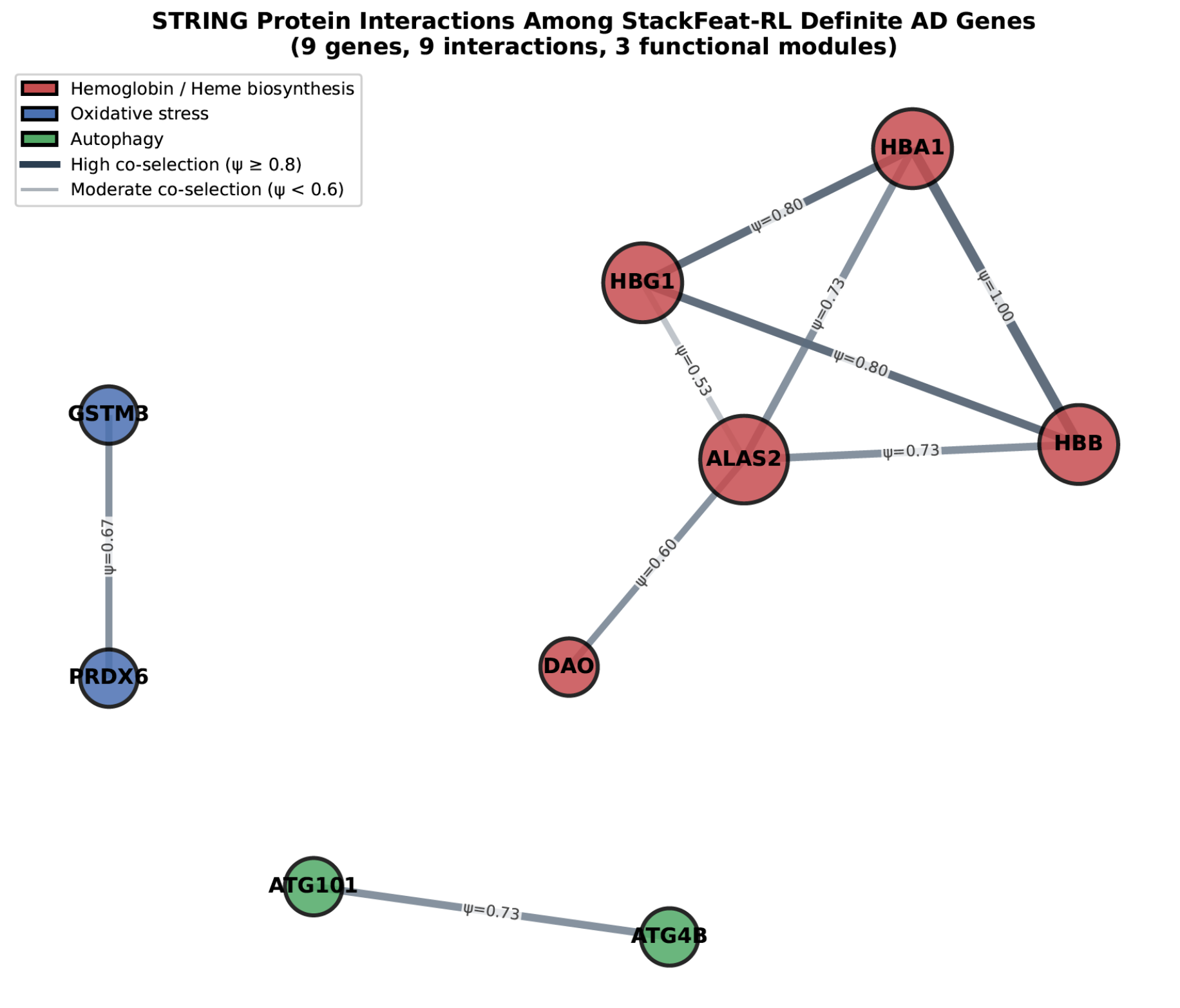}
		\caption{STRING protein interactions among StackFeat-RL Definite AD
			genes filtered by co-selection frequency ($\psi$). Edge labels
			show $\psi$ values. Three functional modules emerge:
			haemoglobin/haeme biosynthesis (red), oxidative stress (blue), and
			autophagy (green). Node size proportional to degree.}
		\label{fig:network}
	\end{figure}
	
	\section{Discussion}
	
	StackFeat-RL achieves Pareto-dominant performance on the accuracy-sparsity frontier, largely due to the following design choices:
	
	\textbf{Dual-criterion accumulation with convergence guarantees.}
	The intersection of features ranked by both coefficient magnitude
	and selection frequency guards against two failure modes that
	single-criterion methods miss: sign-inconsistent features (high
	$c_j$, low $|w_j|$) that would pass stability selection's
	frequency-only threshold, and infrequent-but-consistent features
	(low $c_j$, high $|w_j|$) that would be missed by any
	frequency-based method. Iterative accumulation provides convergence
	to population-level importance measures via the law of large numbers,
	a guarantee absent from Boruta, mRMR, or single-pass elastic net.
	
	\textbf{ElasticNetCV-anchored regularisation.} A single ElasticNetCV
	call on the training data determines $\alpha$, replacing per-iteration
	nested CV and reducing computation by 10--17$\times$. Since
	REINFORCE episodes are independent, they can run in parallel,
	further reducing wall-clock time. This makes StackFeat-RL
	practical on high-dimensional data ($p > 10\,000$) where
	StackFeat would be prohibitively expensive.
	
	\textbf{RL-optimised hyperparameters.} The REINFORCE policy adjusts the feature retention fraction
	m\_frac from the initial 0.575 to 0.49--0.56 across datasets,
	consistently maintaining a wider candidate pool than the StackFeat
	default of 0.25. This allows the dual criterion to evaluate more
	candidates before intersection, reducing the risk of prematurely
	filtering informative genes. The per-gene components
	($\theta_1$--$\theta_4$) converge to zero on all tested datasets,
	which is itself a finding: the policy learns that uniform penalties
	are sufficient for these phenotypes, and no per-gene modulation is
	needed. The framework supports arbitrary per-gene priors (network
	topology, pathway membership, multi-omics concordance) through the
	state representation, and future datasets with stronger biological
	structure may activate these components. 
	Combined with ElasticNetCV-anchored $\alpha$ and convergence-based
	stopping, this makes StackFeat-RL fully automatic: no manual
	specification of panel size, regularisation strength, or stopping
	criterion is needed.
	
	\textbf{Biological output.} Beyond gene lists, StackFeat-RL produces
	a posterior interaction network $\mathbf{M}^*$ that filters the
	STRING interactome to disease-relevant edges via co-selection
	evidence. On Definite AD, this reveals three functional modules
	(haemoglobin/haeme biosynthesis, oxidative stress, autophagy) that
	are consistent with established AD pathobiology and independently
	corroborated by pathway enrichment analysis.
	
	\textbf{Limitations.} The Bernoulli independence assumption is a
	standard approximation; elastic net selections are in practice
	correlated. Evaluation covers two datasets (one miRNA, one
	microarray); validation on RNA-seq, cancer subtyping, and
	non-genomic high-dimensional data would strengthen generality.
	
	\textbf{Future work.} The per-gene policy components
	($\theta_1$--$\theta_4$) did not activate on the current datasets,
	suggesting that datasets with stronger network structure (e.g.,
	cancer with known driver pathways) may be needed to exploit
	per-gene modulation. Richer policy architectures (e.g., neural
	networks with attention over genes) could capture gene-gene
	interactions in the penalty structure. Since REINFORCE episodes
	are independent, they can be trivially parallelised across cores,
	making StackFeat-RL scalable to genome-wide data ($d > 50\,000$)
	with near-linear speedup. Multi-cohort validation with external
	test sets would assess clinical translatability of the selected
	panels.
	
	\section{Competing interests}
	No competing interest is declared.
	
	\section{Author contributions statement}
	
	A.Y.\ conceived the dual-criterion algorithm, designed and
	implemented StackFeat-RL, conducted the experiments, and wrote
	the manuscript. D.A.H.M.\ proposed the convergence-based stopping
	criterion (StackFeat 1.0), introduced the REINFORCE formulation
	and the use of biological priors, and provided critical
	methodological feedback throughout development.
	
	\section{Acknowledgments}
	
	The miRNA expression
	data (GSE240888) and Alzheimer's disease data (GSE84422) are publicly
	available from the NCBI Gene Expression Omnibus.
	
	\section*{Funding}
	
	This work was supported by PAfoS.AI.

\end{document}